\documentclass[letterpaper, 10 pt, conference]{ieeeconf}

\IEEEoverridecommandlockouts
\usepackage{cite}
\usepackage{amsmath,amssymb,amsfonts}
\usepackage{algorithmic}
\usepackage{graphicx}
\usepackage{textcomp}
\usepackage{xcolor}

\usepackage{booktabs}
\usepackage{multirow}
\usepackage{amssymb}
\usepackage{textcomp}
\usepackage{makecell}

\usepackage{graphicx}
\usepackage{capt-of}

\usepackage[table]{xcolor}
\definecolor{bestcol}{RGB}{255,200,200}    
\definecolor{secondcol}{RGB}{255,236,145}   
\newcommand{\best}[1]{\cellcolor{bestcol}#1}
\newcommand{\second}[1]{\cellcolor{secondcol}#1}

\title{\LARGE \bf SFE-VGGT: Source-Free VGGT Distillation for Event-Based Monocular Depth Estimation}

\author{
Thai Duy Nguyen$^{1}$ and Addison Lin Wang$^{1,*}$%
\thanks{$^{1}$Nanyang Technological University, Singapore.
{\tt\small nguyendu003@e.ntu.edu.sg, linwang@ntu.edu.sg}}%
\thanks{$^{*}$Corresponding author.}%
}

\begin{document}

\IEEEaftertitletext{
    \begin{center}
        \includegraphics[width=\textwidth]{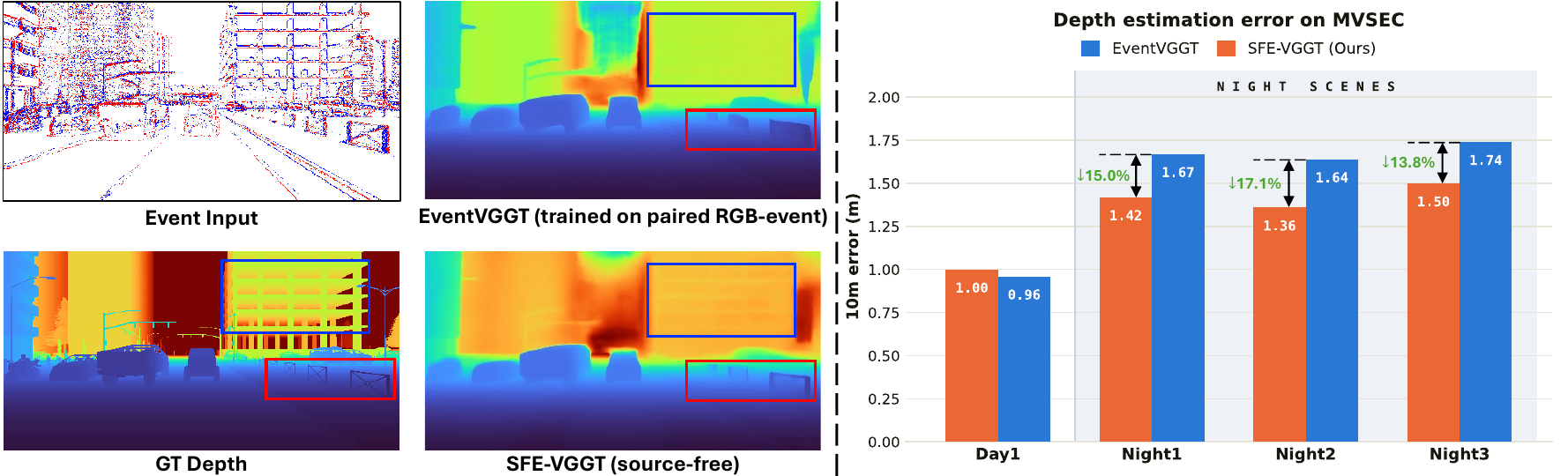}
        \vspace{-12pt}
        \captionof{figure}{
        Our method (SFE-VGGT) distills the geometric priors of VGGT to the event domain in a source-free manner.
        Despite requiring no paired RGB observations, our
        method achieves performance comparable to EventVGGT, which is trained
        with paired RGB-event data.
        Under challenging nighttime conditions on MVSEC, our
        method achieves lower 10\,m depth error, demonstrating
        that effective transfer of VGGT's geometric priors to the event domain
        is possible without paired RGB supervision.
        (\textbf{Left}: Qualitative results on EventScape dataset. \textbf{Right}: Quantiative results on real-world MVSEC dataset.)
        }
        \label{fig:teaser}
    \end{center}
}

\maketitle
\thispagestyle{empty}
\pagestyle{empty}


\begin{abstract}
Recent event-based depth estimation methods successfully transfer geometric priors from vision foundation models via cross-modal distillation. However, their reliance on synchronized RGB-event pairs or depth annotations during training severely restricts practical deployment. To overcome this bottleneck, we propose SFE-VGGT, a novel source-free framework that distills the geometric priors of VGGT to the event domain without any paired RGB observations. Our \textit{core idea} is to reconstruct surrogate frames directly from the target event stream to act as a frozen geometric teacher, entirely eliminating the need for genuine source RGB data. Crucially, as these surrogate frames inherently yield imperfect and spatially varying supervision, directly distilling from them propagates artifacts. To resolve this, we introduce a novel reliability-aware distillation strategy. This includes Density-Aware Feature Distillation to emphasize informative event regions, and Confidence-Weighted Depth Distillation to dynamically regulate supervision based on relative teacher-student prediction confidence. Meanwhile, we propose a Cross-Frame Relational Consistency loss that enforces temporal geometric stability using reliable inter-frame correspondences, bypassing the need for temporally-consistent teacher's depth. Extensive experiments demonstrate that, despite source-free, our SFE-VGGT closely matches the accuracy of RGB-dependent baselines under standard conditions, and significantly surpasses them in challenging nighttime scenarios. Across MVSEC nighttime sequences, SFE-VGGT reduces the average 10\,m depth error by 15.3\% compared with EventVGGT. Moreover, our method exhibits robust zero-shot generalization across real-world datasets, proving that highly effective geometric priors can be transferred to event cameras using strictly source-free supervision.
\end{abstract}
\section{Introduction}
\label{sec:introduction}

Event cameras asynchronously measure changes in scene brightness, providing high temporal resolution, high dynamic range, and low latency~\cite{gallego2022event}. These properties make them particularly attractive for geometric perception in scenarios involving rapid motion or challenging illumination, where conventional frame-based cameras typically suffer from motion blur, saturation, or insufficient exposure. Among event-based perception tasks, monocular depth estimation is critical for applications like autonomous navigation and robotic perception. However, learning dense depth directly from events remains difficult because event streams are sparse and fundamentally different from images.
Such sparsity and spatial variation are also encountered in prior
event-based depth methods~\cite{srfnet,depthanyevent,eventdam}.
Moreover, large-scale datasets with dense metric depth annotations are
expensive and scarce, constraining early supervised approaches~\cite{e2depth}
and event-image fusion methods~\cite{ramnet,srfnet,erf2d} to
task-specific datasets.

Recent vision foundation models provide a compelling alternative to direct supervision. Large-scale image-based models encapsulate strong semantic and geometric priors that can be transferred to event representations through cross-modal knowledge distillation. Methods such as Depth AnyEvent~\cite{depthanyevent} and EventDAM~\cite{eventdam} successfully exploit image-based depth foundation models to supervise event-based students without requiring metric depth annotations. More recently, EventVGGT~\cite{eventvggt} demonstrated that the powerful multi-view geometric priors of VGGT~\cite{vggt} can be transferred to temporally ordered event sequences, substantially improving both depth accuracy and temporal consistency.

Crucially, however, eliminating the need for depth annotations does not eliminate the dependence on the original image modality. Existing distillation-based approaches strictly assume that synchronized or paired RGB observations are available alongside the event stream during training~\cite{eventdam,depthanyevent,eventvggt}. This assumption severely restricts practical deployment, as it necessitates a multi-modal acquisition setup with calibrated and temporally aligned cameras. Consequently, training is confined to datasets where the source image modality was recorded. This critical limitation raises a fundamental research question: \emph{can the geometric priors of an image-based foundation model like VGGT be distilled into an event-based model using strictly source-free supervision, without any access to the original RGB observations?}

To this end, we propose a novel source-free distillation framework, called \textbf{SFE-VGGT}. Our \textit{core idea} is to reconstruct surrogate frames directly from the target event stream to act as a frozen geometric teacher, entirely eliminating the need for genuine source RGB data. Specifically, we utilize a frozen recurrent event-to-video reconstruction network~\cite{e2vid} to convert events into surrogate frames. These frames are subsequently processed by a frozen VGGT~\cite{vggt}, exposing its pretrained geometric priors purely from event-derived data. Concurrently, a student initialized from the same VGGT backbone processes frame-like event representations, adapting via lightweight trainable parameters. At inference time, only the student model is retained.

However, this source-free formulation introduces a critical technical challenge: \textit{event reconstruction is inherently imperfect, yielding spatially varying and sometimes unreliable supervision.} Directly forcing a student network to reproduce all teacher features and depths from these reconstructed frames inevitably propagates artifacts. To resolve this, we introduce a \textit{reliability-aware distillation strategy} comprising three main technical contributions. First, \emph{Density-Aware Feature Distillation (\textbf{DAFD})} weights feature alignment according to patch-level event activity, ensuring that regions lacking sufficient event evidence do not dominate the transfer process. Second, \emph{Confidence-Weighted Depth Distillation (\textbf{CWDD})} performs scale-invariant depth transfer while dynamically regulating pixel-wise supervision based on the relative confidence between the teacher and student. Finally, we formulate a \emph{Cross-Frame Relational Consistency (\textbf{CFRC})} loss that leverages reliable inter-frame correspondences from the reconstructed frames to preserve the student's relative depth ordering across time, bypassing the need to directly distill temporal changes from potentially noisy teacher depth maps.

Extensive experiments on the EventScape \cite{eventscape}, MVSEC  \cite{mvsec}, and DENSE \cite{dense} datasets demonstrate the effectiveness of this approach. Our findings show that the proposed framework successfully retains the formidable geometric capabilities of VGGT despite operating in a restrictive, source-free setting. Under standard evaluation conditions, SFE-VGGT closely matches the performance of the RGB-dependent EventVGGT baseline. More notably, as depicted in Fig.~\ref{fig:teaser}, our source-free approach significantly outperforms EventVGGT on challenging nighttime MVSEC sequences, where the inherent advantages of event sensing are most pronounced. In particular, SFE-VGGT reduces the average 10\,m and 20\,m depth errors by 15.3\% and 4.0\%, respectively. Moreover, the model exhibits robust zero-shot generalization from synthetic EventScape data to unseen real-world sequences, decisively proving that access to the original RGB modality is not a strict prerequisite for transferring highly effective geometric priors to event cameras.

In summary, our main contributions are four-fold: (\textbf{I}) We introduce SFE-VGGT, a novel source-free distillation framework for event-based monocular depth estimation that eliminates the need for paired RGB data or metric depth labels. (\textbf{II})  We propose a reliability-aware distillation strategy (including DAFD and CWDD) to dynamically suppress uncertain regions arising from imperfect surrogate frames. (\textbf{III}) We propose a CFRC loss that enforces temporal depth stability using reliable reconstructed-frame correspondences, without relying on absolute VGGT depth maps. (\textbf{IV}) Our source-free method achieves on-par performance with the RGB-dependent methods and significantly better performance in nighttime scenarios.

\begin{figure*}[t]
    \centering
    \includegraphics[width=\textwidth]{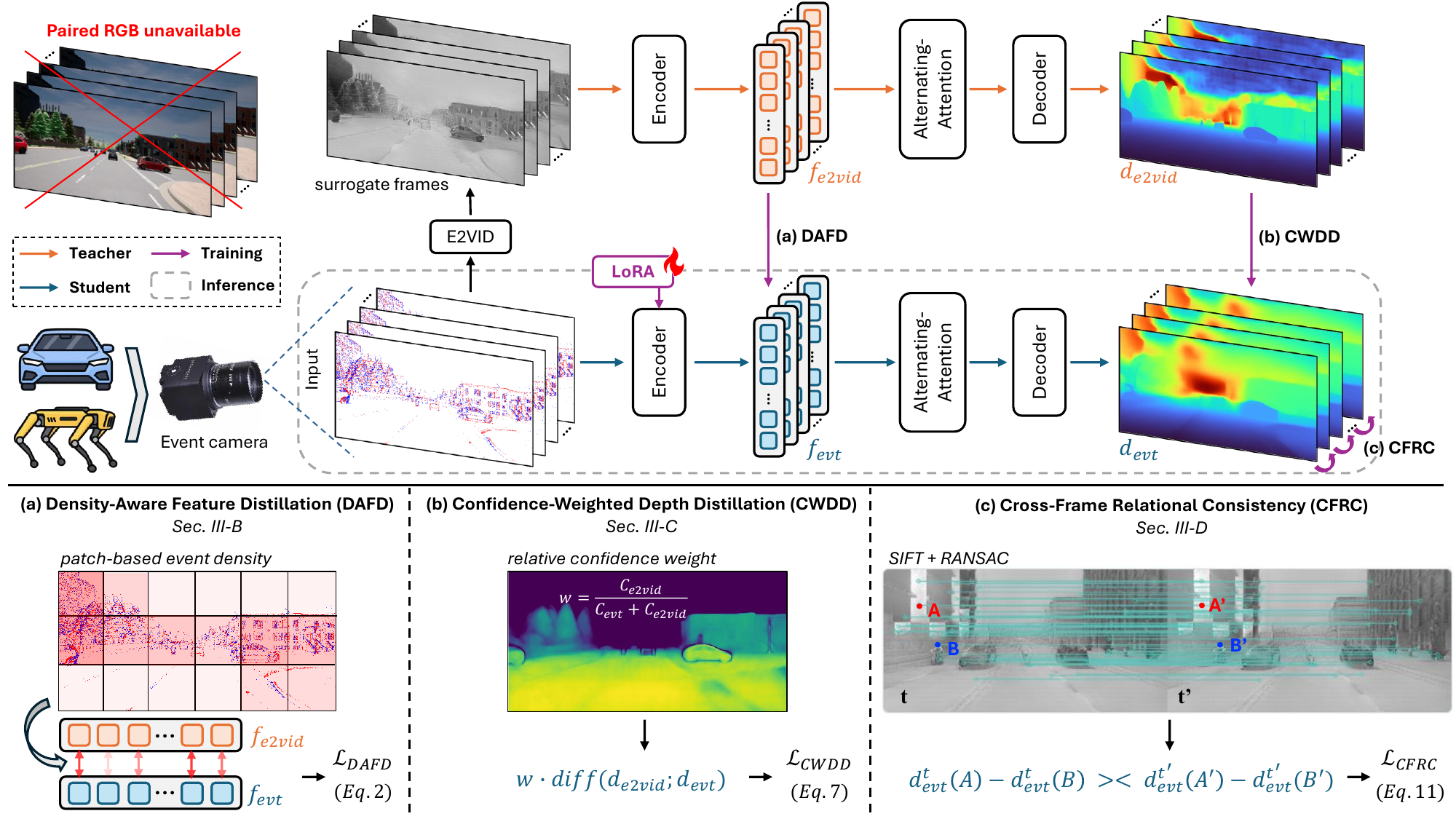}
    \caption{
    \textbf{Overview of the proposed SFE-VGGT framework.}
    During training, the event data is processed by a pretrained E2VID model to construct surrogate frames for the frozen VGGT teacher, while the student directly processes event representations. Knowledge is transferred through (a) Density-Aware Feature Distillation (DAFD), which weights patch-level feature alignment by event density, (b) Confidence-Weighted Depth Distillation (CWDD), which weights depth supervision using relative prediction confidence, and (c) Cross-Frame Relational Consistency (CFRC), which enforces consistent depth relations across matched points between frames. During inference, only the event-based student is retained.
    }
    \label{fig:methodology}
    \vspace{-12pt}
\end{figure*}

\section{Related Works}
\label{sec:related_work}

\paragraph{Vision foundation models}
Large-scale pretraining has produced highly transferable visual
representations through image-text supervision, self-supervised
learning, and masked modeling
~\cite{clip,dino,mae,dinov2}.
Foundation models have subsequently demonstrated strong generalization
for segmentation and monocular geometry
~\cite{sam,depthanythingv2}, and their priors are increasingly being
adapted to event-based perception
~\cite{eventsam,openess,depthanyevent,eventdam}.
Among geometric foundation models, VGGT~\cite{vggt} jointly reasons
about depth, cameras, point maps, and tracks across multiple views.
EventVGGT~\cite{eventvggt} shows that these multi-view priors can be
transferred to event sequences when synchronized RGB data are
available.
\textit{Our work considers the complementary source-free setting, where the
teacher observation is reconstructed solely from events and used for
reliability-aware geometric distillation.}

\paragraph{Event-based monocular depth estimation}
Event cameras provide high temporal resolution and dynamic range,
making them attractive for geometric perception under fast motion and
challenging illumination~\cite{gallego2022event}.
Early event-only methods estimate dense depth using recurrent or
transformer-based architectures
~\cite{e2depth,evt,ereformer,even}, while another line of work combines
events with conventional frames to exploit complementary temporal and
photometric information
~\cite{ramnet,hmnet,srfnet,erf2d,cfrnet}.
Despite their strong performance, supervised approaches are limited by
the scarcity of dense event-depth annotations.
Recent methods therefore transfer knowledge from pretrained image
models~\cite{depthanyevent,eventdam,eventvggt}.
Depth AnyEvent~\cite{depthanyevent} and EventDAM~\cite{eventdam}
distill image-domain depth priors into event representations, while
EventVGGT~\cite{eventvggt} further transfers the geometric
priors of VGGT to temporally ordered event sequences.
\textit{However, these approaches still rely on paired or synchronized RGB
observations during distillation.
Our method instead derives the teacher view directly from the event
stream.}

\paragraph{Source-free knowledge distillation}
Knowledge distillation transfers predictions or representations from
a pretrained teacher to a student
~\cite{hinton2015distilling,wang2021knowledge}, and has been extended across
diverse modalities for representation learning, temporal
understanding, and 3D perception
~\cite{gupta2016crossmodal,lee2023decomposed,stxd}.
Source-free distillation further removes access to the original
task-relevant source data
~\cite{shot,sfda_neighbors,sfdade}, with recent works extending this
setting to cross-modal and event-based transfer under limited or
unavailable source observations
~\cite{socket,sourcefreecmkt,eventdance}.
These approaches aim to preserve transferable knowledge despite the
absence of the original source modality, while addressing the
distribution and representation gaps between heterogeneous sensing
domains.
\textit{However, unlike prior works that primarily focus on recognition tasks, our method targets dense geometric prediction by distilling the geometric priors of a frozen VGGT teacher to an event-based depth student.}

\section{Methodology}
\label{sec:method}


\subsection{Overview and Problem Formulation}
\label{sec:overview}

\noindent \textbf{Framework overview.}
We propose a source-free framework for event-based depth estimation that transfers geometric priors from a pretrained VGGT model~\cite{vggt} to the event domain without paired RGB observations. As depicted in Fig.~\ref{fig:methodology}, our method reconstructs intensity frames from the event stream using a pretrained E2VID model~\cite{e2vid} and feeds them to the teacher branch, while the student directly processes event representations.
Both branches are initialized from the same pretrained VGGT checkpoint,
and no paired RGB data are required.
Both branches share the same pretrained VGGT initialization, and
training requires no paired RGB data.

Since the teacher operates on reconstructed frames, its supervision can be affected by reconstruction artifacts and spatial uncertainty. Directly enforcing uniform agreement may propagate unreliable teacher signals to the student. To make the transfer more selective, we
use three complementary objectives: Density-Aware Feature
Distillation (Sec. \ref{sec:feat_loss}), Confidence-Weighted Depth Distillation (Sec. \ref{sec:depth_loss}), and Cross-Frame Relational Consistency (Sec. \ref{sec:rank_loss}).

\smallskip
\noindent \textbf{Event-based Monocular Depth Estimation.}
Event cameras encode visual changes as a stream of asynchronous events.
Each event is represented as
$e_i=(x_i,y_i,t_i,p_i)$, where $(x_i,y_i)$ denotes the pixel location,
$t_i$ is the timestamp, and $p_i\in\{-1,+1\}$ indicates the polarity.
To enable processing by the frame-based VGGT architecture, we partition
the continuous event stream into $S$ consecutive temporal windows and
accumulate them into a sequence of event representations
$\mathbf{I}^{\mathrm{evt}}
=\{I_s^{\mathrm{evt}}\}_{s=1}^{S}$,
where $I_s^{\mathrm{evt}}\in\mathbb{R}^{3\times H\times W}$.
The resulting sequence is used as input to the event-based depth model.

Following VGGT~\cite{vggt}, each event frame is partitioned into
non-overlapping patches and embedded into patch tokens.
The tokens from the $S$ frames are processed by the VGGT backbone,
yielding event features
$f_{\mathrm{evt}}\in\mathbb{R}^{S\times K\times C}$,
where $K$ and $C$ denote the number of patch tokens and feature
dimension, respectively.
The prediction head outputs per-frame depth prediction $d_{\mathrm{evt}}$ and
confidence $C_{\mathrm{evt}}$.

\subsection{Density-Aware Feature Distillation (DAFD)}
\label{sec:feat_loss}

\noindent \textbf{Insight:} Event observations are inherently sparse and spatially non-uniform,
such that different regions provide substantially different amounts of
visual evidence. This affects both sides of our distillation framework.
For the student, features extracted from event-active regions are better
supported by the input than those from weakly observed patches.
Likewise, the surrogate frames used by the teacher are
also less reliable in regions where insufficient events are available
for reconstruction.
Therefore, \textbf{uniformly aligning all feature tokens may overemphasize
regions where either branch is weakly supported by the observed events.}

To account for this, we use the local event density as a reliability
signal for feature distillation. As shown in Fig.\ref{fig:methodology}(a), for the $i$-th patch of size $P\times P$ at timestep $s$, we define
\begin{equation}
    w_{s,i}=\frac{N_{s,i}}{P\times P},
\end{equation}
where $N_{s,i}$ denotes the number of spatial locations activated by
at least one event. Hence, $w_{s,i}\in[0,1]$ measures the event support of the
corresponding patch token. We directly associate this weight with the
feature token extracted from the same spatial patch, such that a denser
patch receives stronger feature-level supervision, while a sparse patch
contributes less to the distillation objective.
Given student and teacher features
$f_{\mathrm{evt}},f_{\mathrm{e2vid}}
\in\mathbb{R}^{S\times K_f\times C}$,
we define the density-aware feature distillation loss as
\begin{equation}
    \mathcal{L}_{\mathrm{DAFD}}
    =
    \frac{
        \sum_{s,i} w_{s,i}
        \left[
        1-\cos\!\left(
            f_{\mathrm{evt}}^{s,i},
            \operatorname{sg}[f_{\mathrm{e2vid}}^{s,i}]
        \right)
        \right]
    }{
        \sum_{s,i} w_{s,i}+\epsilon
    },
\end{equation}
where $\operatorname{sg}[\cdot]$ denotes stop-gradient.
The density weighting places greater emphasis on regions where both
the event representation and the reconstructed teacher observation are
better supported by the events, while reducing the influence
of potentially unreliable alignment in sparse regions. 
Normalizing by the accumulated density prevents variations in
event activity from affecting the scale of the objective.

\subsection{Confidence-Weighted Depth Distillation (CWDD)}
\label{sec:depth_loss}

\noindent \textbf{Insight:} Feature alignment alone does not directly constrain the predicted
geometry. We therefore further distill the teacher depth into the event
student. However, the teacher prediction is obtained from E2VID
frames may inherit errors from uncertain reconstruction regions. \textbf{Uniform depth supervision could therefore force the student to reproduce unreliable teacher predictions.} Since both branches jointly predict depth and confidence, we use their relative confidence to adapt the supervision strength at
each pixel.

As shown in Fig.\ref{fig:methodology}(b), for each pixel $i$, we define the confidence weight as
\begin{equation}
    w_i =
    \frac{
        \operatorname{sg}[C_{\mathrm{e2vid},i}]
    }{
        \operatorname{sg}[C_{\mathrm{evt},i}]
        +
        \operatorname{sg}[C_{\mathrm{e2vid},i}]
        + \epsilon
    },
    \qquad
    \tilde{w}_i =
    \frac{w_i}{\operatorname{mean}_j(w_j)},
\end{equation}
where $\operatorname{sg}[\cdot]$ denotes stop-gradient.
The ratio reflects the teacher confidence relative to that of the
student. When
$C_{\mathrm{e2vid}}\gg C_{\mathrm{evt}}$, $w_i\rightarrow1$ and the
teacher provides stronger supervision; when
$C_{\mathrm{evt}}\gg C_{\mathrm{e2vid}}$, $w_i\rightarrow0$ and the
teacher contribution is suppressed. Comparable confidences yield
$w_i\approx0.5$. 
The bounded ratio prevents high-confidence predictions from dominating
the loss, while mean normalization stabilizes its magnitude across samples.

Since monocular depth is ambiguous up to a global scale, we perform
distillation in log-depth space using a scale-invariant
objective~\cite{eigen2014depth}. Let
\begin{equation}
    r_i =
    \log d_{\mathrm{evt},i}
    -
    \operatorname{sg}\!\left[\log d_{\mathrm{e2vid},i}\right].
\end{equation}
The confidence-weighted scale-invariant term is
\begin{equation}
    \mathcal{L}_{\mathrm{si}}
    =
    \mathbb{E}_i\!\left[\tilde{w}_i r_i^2\right]
    -
    \lambda_{\mathrm{si}}
    \left(
        \mathbb{E}_i\!\left[\tilde{w}_i r_i\right]
    \right)^2 .
\end{equation}
To additionally preserve local depth structure, we introduce a
gradient consistency term,
\begin{equation}
    \mathcal{L}_{\mathrm{grad}}
    =
    \frac{1}{2}
    \mathbb{E}_i
    \left[
        \tilde{w}_i
        \left(
            |\nabla_x r_i|
            +
            |\nabla_y r_i|
        \right)
    \right].
\end{equation}
The final depth distillation objective is
\begin{equation}
    \mathcal{L}_{\mathrm{CWDD}}
    =
    \mathcal{L}_{\mathrm{si}}
    +
    \gamma\,\mathcal{L}_{\mathrm{grad}}.
\end{equation}
This objective transfers both relative depth values and local geometric
structure while reducing the influence of uncertain teacher
predictions.

\subsection{Cross-Frame Relational Consistency (CFRC)}
\label{sec:rank_loss}

\noindent \textbf{Insight:} The preceding objectives supervise feature and depth predictions but do
not \textbf{explicitly constrain temporal relations.} 
We therefore use the reconstructed frames to establish reliable
cross-frame correspondences, while deriving relational supervision from
the student's depth predictions.

As depicted in Fig.\ref{fig:methodology}(c), for a frame pair $(t,t')$, SIFT features~\cite{lowe2004distinctive}
are matched on the reconstructed frames and geometric outliers are
removed using RANSAC~\cite{fischler1981random}. Let
$(u_j^t,u_j^{t'})$ denote a valid correspondence. From these matches,
we form pairs of correspondence tracks $(a_k,b_k)$ and use the
student's depth ordering in frame $t$ as a detached reference:
\begin{equation}
    r_k =
    \operatorname{sign}
    \left(
        \operatorname{sg}\!\left[
        d_{\mathrm{evt}}^{t}(u_{a_k}^{t})
        -
        d_{\mathrm{evt}}^{t}(u_{b_k}^{t})
        \right]
    \right).
\end{equation}
The corresponding depth difference in frame $t'$ is
\begin{equation}
    \Delta_k =
    d_{\mathrm{evt}}^{t'}(u_{a_k}^{t'})
    -
    d_{\mathrm{evt}}^{t'}(u_{b_k}^{t'}).
\end{equation}

\noindent For each relational pair, we define
\begin{equation}
    \ell_k =
    \mathbf{1}[r_k\neq0]\,
    \operatorname{softplus}(-r_k\Delta_k)
    +
    \mathbf{1}[r_k=0]\,\Delta_k^2 .
\end{equation}
Let $\mathcal{T}$ denote the set of frame pairs and
$\mathcal{K}_{t,t'}$ the relational pairs associated with
$(t,t')$. The final objective is
\begin{equation}
    \mathcal{L}_{\mathrm{CFRC}}
    =
    \frac{1}{|\mathcal{T}|}
    \sum_{(t,t')\in\mathcal{T}}
    \frac{1}{|\mathcal{K}_{t,t'}|}
    \sum_{k\in\mathcal{K}_{t,t'}}
    \ell_k .
\end{equation}

For $r_k=+1$ or $-1$, the loss encourages the same relative depth
ordering to be preserved in frame $t'$, while the zero case preserves
a near-equal relation. The loss is first averaged within each frame
pair and then across frame pairs, preventing pairs with more matches
from dominating the objective. Notably, the teacher
provides only correspondence geometry; no teacher depth values are
used in this loss. 

\noindent \textbf{Total objective.}
The final training objective combines the three proposed objectives as
\begin{equation}
    \mathcal{L}
    =
    \lambda_{\mathrm{DAFD}}\mathcal{L}_{\mathrm{DAFD}}
    +
    \lambda_{\mathrm{CWDD}}\mathcal{L}_{\mathrm{CWDD}}
    +
    \lambda_{\mathrm{CFRC}}\mathcal{L}_{\mathrm{CFRC}},
\end{equation}
where $\lambda_{\mathrm{DAFD}}$, $\lambda_{\mathrm{CWDD}}$, and
$\lambda_{\mathrm{CFRC}}$ balance feature-level transfer,
depth-level geometric supervision, and cross-frame relational
consistency, respectively. Empirically, we set
$\lambda_{\mathrm{DAFD}}=1.0$,
$\lambda_{\mathrm{CWDD}}=2.0$, and
$\lambda_{\mathrm{CFRC}}=0.2$.

\section{Experiments}
\label{sec:experiments}

\subsection{Experimental Settings}

\noindent \textbf{Datasets.}
Following the evaluation protocol of prior works~\cite{eventvggt,eventdam,srfnet},
we use EventScape~\cite{eventscape} as the primary training dataset due to its large scale. Subsequently, its test set is used for in-domain evaluation. For zero-shot evaluation, models trained on EventScape are directly tested on the unseen MVSEC~\cite{mvsec} and DENSE~\cite{dense} datasets without any finetuning or adaptation.

\smallskip
\noindent \textbf{Evaluation metrics.}
We report mean absolute depth error at mutiple cut-off ranges of 10\,m, 20\,m,
and 30\,m. Additionally, percentage metrics $\delta_i$ at
$i\in\{1.25,1.25^2,1.25^3\}$ are reported where applicable.

\smallskip
\noindent \textbf{Implementation details.}
Following prior works~\cite{eventvggt,eventdam,srfnet}, event and E2VID-reconstructed inputs are
center-cropped to $(252\times504)$.
For event reconstruction, we use the publicly released pretrained
E2VID model~\cite{e2vid}.
Aligning with the VGGT~\cite{vggt} training protocol, sky regions are masked out to exclude invalid depth values. For comparison with EventVGGT~\cite{eventvggt}, we adapt the frozen VGGT backbone using a similar LoRA configuration ($r=16$, $\alpha=32$),
resulting in approximately 1.6 million trainable parameters. The model
is optimized with AdamW at a learning rate of $1\times10^{-5}$.
Training uses 24-frame windows on a single NVIDIA RTX 5090, converging in roughly 23 hours.

\subsection{Experiment Results}
\label{sec:exp_results}

\subsubsection{Results on EventScape dataset}
\label{sec:eventscape_results}

As shown in Tab.~\ref{tab:eventscape}, SFE-VGGT remains highly
competitive across all evaluation ranges despite operating in a
source-free setting. \textbf{At 10\,m, it achieves comparable performance to
EventVGGT}~\cite{eventvggt}, while
remaining close to the strongest supervised and distillation-based
baselines. The advantage becomes more evident as the evaluation range
increases. SFE-VGGT achieves the \textbf{second-best overall performance
at both 20\,m and 30\,m}, outperforming all supervised event-image
methods at these two ranges. Compared with EventDAM~\cite{eventdam},
the errors are reduced by \textbf{42.8\%} and \textbf{46.5\%} at 20\,m and 30\,m,
respectively. It also improves over the strongest supervised event-image
baseline (SRFNet~\cite{srfnet}) at these ranges by \textbf{48.2\%} and \textbf{55.4\%}, showing that the
transferred VGGT priors remain effective for medium- and long-range
geometry. Although EventVGGT retains the best overall accuracy,
\textbf{SFE-VGGT achieves competitive depth estimation without requiring
paired RGB observations during distillation}. This result is particularly
noteworthy because SFE-VGGT is the only source-free approach among the
compared distillation methods, demonstrating that strong geometric
transfer can be retained even when the original RGB modality is
unavailable. Qualitative results are shown in Fig.~\ref{fig:teaser}.

\begin{table}[t!]
    \centering
    \caption{
    Mean absolute depth error on EventScape in meters.\\
    E denotes event-only input and E+I denotes event and RGB image input.
    ``Depth GT'' indicates whether metric ground-truth depth values
    are used as training supervision.
    \colorbox{bestcol}{Best} and \colorbox{secondcol}{second-best} results are highlighted.
    }
     \vspace{-10pt}
    \label{tab:eventscape}
    \scriptsize
    \setlength{\tabcolsep}{2.3pt}
    \begin{tabular*}{\columnwidth}{@{\extracolsep{\fill}}lccc|ccc@{}}
        \toprule
        \multirow{2}{*}{Method}
        & \multicolumn{2}{c}{Training}
        & \multicolumn{1}{c|}{Inference}
        & \multirow{2}{*}{10m $\downarrow$}
        & \multirow{2}{*}{20m $\downarrow$}
        & \multirow{2}{*}{30m $\downarrow$} \\
        \cmidrule(lr){2-3}
        \cmidrule(lr){4-4}
        & Source-free & Depth GT & Input & & & \\
        \midrule

        E2Depth~\cite{e2depth}
        & -- & \checkmark & E
        & 1.79 & 5.35 & 8.31 \\

        \midrule

        RAMNet~\cite{ramnet}
        & -- & \checkmark & E+I
        & 0.81 & 2.26 & 3.58 \\

        HMNet~\cite{hmnet}
        & -- & \checkmark & E+I
        & \second{0.55} & 1.80 & 3.27 \\

        ER-F2D~\cite{erf2d}
        & -- & \checkmark & E+I
        & 0.67 & 1.69 & 2.81 \\

        SRFNet~\cite{srfnet}
        & -- & \checkmark & E+I
        & 1.27 & 1.68 & 2.76 \\

        \midrule

        EventDAM~\cite{eventdam}
        & \texttimes & \texttimes & E
        & 0.56 & 1.52 & 2.30 \\

        EventVGGT~\cite{eventvggt}
        & \texttimes & \texttimes & E
        & \best{0.54}
        & \best{0.79}
        & \best{1.06} \\
        
        \textbf{SFE-VGGT}
        & \checkmark & \texttimes & E
        & 0.57
        & \second{0.87}
        & \second{1.23} \\

        \bottomrule
    \end{tabular*}
\end{table}

\begin{table*}[t]
\centering
\caption{
Mean absolute depth error on MVSEC in meters. \\
E denotes event-only input and E+I denotes event and RGB image input.
``Depth GT'' indicates whether metric ground-truth depth values are used
as training supervision.
Among distillation-based methods, the \colorbox{bestcol}{Best} and \colorbox{secondcol}{second-best} results are highlighted. Overall best results across all methods are shown in \textbf{bold}.
}
 \vspace{-10pt}
\label{tab:mvsec}
\scriptsize
\setlength{\tabcolsep}{1.8pt}

\begin{tabular*}{\textwidth}{@{\extracolsep{\fill}}lccc|ccc|ccc|ccc|ccc@{}}
\toprule

\multirow{2}{*}{Method}
& \multicolumn{2}{c}{Training}
& \multicolumn{1}{c|}{Inference}
& \multicolumn{3}{c}{Night1}
& \multicolumn{3}{c}{Night2}
& \multicolumn{3}{c}{Night3}
& \multicolumn{3}{c}{Day1} \\

\cmidrule(lr){2-3}
\cmidrule(lr){4-4}
\cmidrule(lr){5-7}
\cmidrule(lr){8-10}
\cmidrule(lr){11-13}
\cmidrule(lr){14-16}

& Source-free & Depth GT & Input
& 10m $\downarrow$ & 20m $\downarrow$ & 30m $\downarrow$
& 10m $\downarrow$ & 20m $\downarrow$ & 30m $\downarrow$
& 10m $\downarrow$ & 20m $\downarrow$ & 30m $\downarrow$
& 10m $\downarrow$ & 20m $\downarrow$ & 30m $\downarrow$ \\

\midrule

E2Depth~\cite{e2depth}
& -- & \checkmark & E
& 3.38 & 3.82 & 4.46
& 1.67 & 2.63 & 3.58
& 1.42 & 2.33 & 3.18
& 1.67 & 2.64 & 3.13 \\

\midrule

RAMNet~\cite{ramnet}
& -- & \checkmark & E+I
& 2.50 & 3.19 & 3.82
& 1.21 & 2.31 & 3.28
& \textbf{1.01} & 2.34 & 3.43
& 1.39 & 2.17 & 2.76 \\

EvT+~\cite{evt}
& -- & \checkmark & E+I
& 1.45 & 2.10 & 2.88
& 1.48 & 2.13 & 2.90
& 1.38 & 2.03 & 2.77
& 1.24 & 1.91 & 2.36 \\

HMNet~\cite{hmnet}
& -- & \checkmark & E+I
& 1.50 & 2.48 & 3.19
& 1.36 & 2.25 & 2.96
& 1.27 & 2.17 & 2.86
& 1.22 & 2.21 & 2.68 \\

ER-F2D~\cite{erf2d}
& -- & \checkmark & E+I
& 1.58 & 2.24 & 2.78
& 1.54 & 2.23 & 2.95
& 1.24 & \textbf{1.96} & 2.81
& 1.34 & 2.25 & 2.62 \\

SRFNet~\cite{srfnet}
& -- & \checkmark & E+I
& \textbf{1.26} & \textbf{1.95} & 3.01
& \textbf{1.19} & 2.13 & 3.22
& \textbf{1.01} & 2.12 & 3.52
& \textbf{0.96} & 1.77 & 2.37 \\

EReFormer~\cite{ereformer}
& -- & \checkmark & E+I
& 1.52 & 2.28 & 2.98
& 1.40 & 2.12 & 2.66
& 1.32 & 2.04 & 2.68
& 1.29 & 2.14 & 2.59 \\

\midrule

EventDAM~\cite{eventdam}
& \texttimes & \texttimes & E
& \best{1.39}
& 2.10
& 3.25
& \second{1.43}
& 2.18
& 3.22
& \best{1.44}
& 2.16
& 3.22
& 1.12
& 1.79
& 2.69 \\

DepthAnyEvent~\cite{depthanyevent}
& \texttimes & \texttimes & E
& 1.87
& 2.27
& \second{2.81}
& 1.99
& 2.40
& 2.86
& 2.05
& 2.49
& 2.97
& 1.50
& 1.97
& 2.40 \\

EventVGGT~\cite{eventvggt}
& \texttimes & \texttimes & E
& 1.67
& \best{2.02}
& \best{\textbf{2.61}}
& 1.64
& \second{2.03}
& \best{\textbf{2.48}}
& 1.74
& \second{2.15}
& \best{\textbf{2.64}}
& \best{\textbf{0.96}}
& \best{\textbf{1.33}}
& \best{\textbf{1.63}} \\

\textbf{SFE-VGGT}
& \checkmark & \texttimes & E
& \second{1.42}
& \second{2.04}
& 2.92
& \best{1.36}
& \best{\textbf{1.90}}
& \second{2.62}
& \second{1.50}
& \best{2.01}
& \second{2.89}
& \second{1.00}
& \second{1.53}
& \second{2.14} \\

\bottomrule
\end{tabular*}
 \vspace{-10pt}
\end{table*}

\begin{figure*}[t!]
    \centering
    \includegraphics[width=0.90\textwidth]{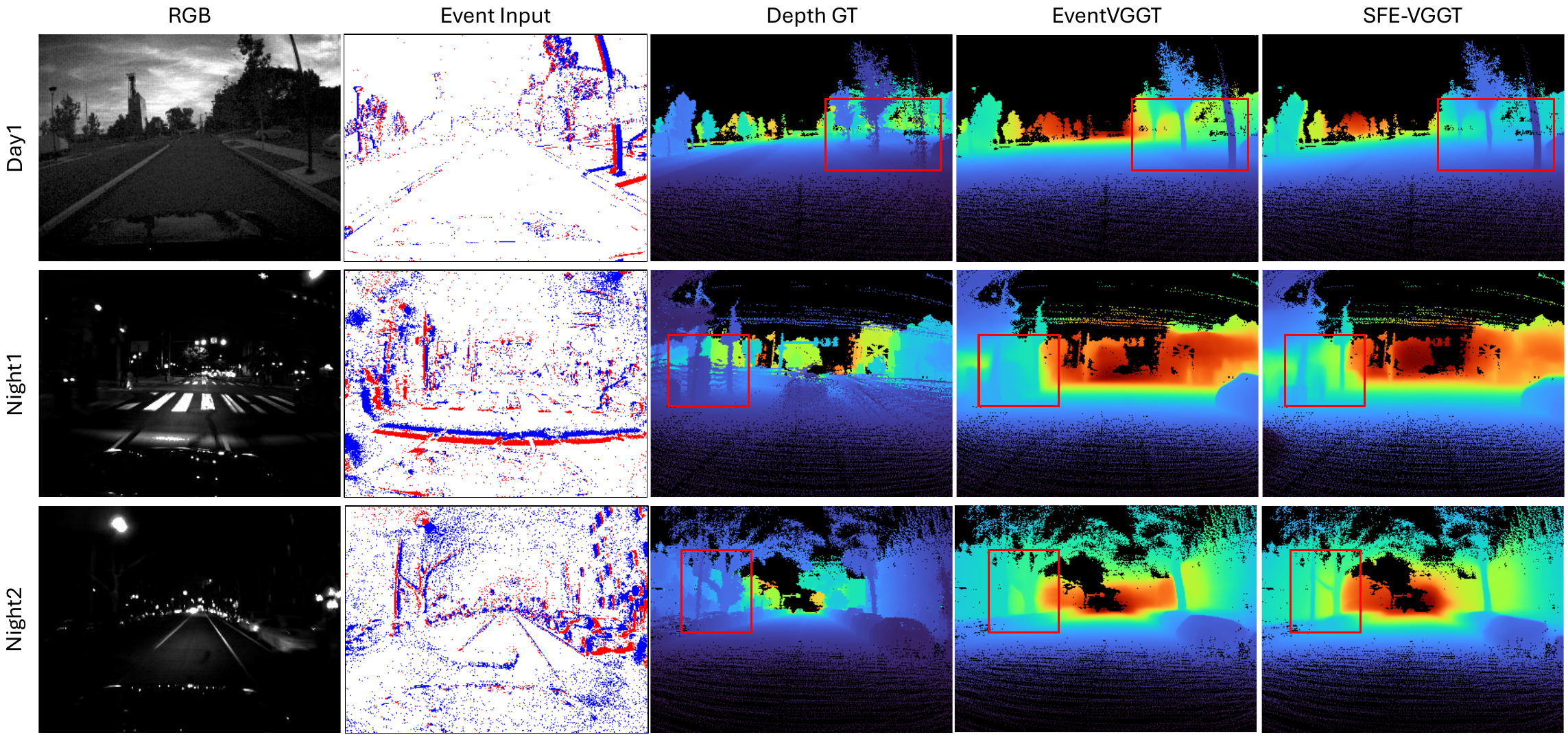}
     \vspace{-10pt}
    \caption{
    Qualitative results on MVSEC dataset.
    }
     \vspace{-15pt}
    \label{fig:mvsec}
\end{figure*}

\subsubsection{Zero-shot results on MVSEC dataset}

The evaluation on MVSEC~\cite{mvsec} examines whether the geometric
priors learned from synthetic EventScape transfer to real-world event
streams without finetuning. Despite the substantial domain shift in
scene appearance, motion, and event statistics, \textbf{SFE-VGGT consistently
ranks among the leading distillation-based methods} across the evaluated
sequences and depth ranges, as shown in Tab.~\ref{tab:mvsec}. Its
advantage is particularly evident under nighttime conditions. At
10\,m, SFE-VGGT consistently outperforms EventVGGT~\cite{eventvggt}
across Night1--Night3 (see Fig.~\ref{fig:teaser}). Averaged over these
three sequences, it reduces the error by \textbf{15.3\%} at 10\,m and
\textbf{4.0\%} at 20\,m relative to EventVGGT, while also achieving the best
overall 20\,m result on Night2. At the farther 30\,m range, EventVGGT
remains stronger, although SFE-VGGT retains second-best performance
among the distillation methods on Night2 and Night3. On Day1,
SFE-VGGT also ranks second among the distillation-based approaches
across all three ranges. Overall, the results demonstrate \textbf{strong
synthetic-to-real generalization}, with the clearest gains appearing in
near- and medium-range nighttime depth estimation. The strong
nighttime performance further suggests that source-free event-based
transfer may be \textbf{particularly beneficial under challenging illumination},
where conventional RGB observations can become degraded. Qualitative
results on Day1, Night1, and Night2 are shown in
Fig.~\ref{fig:mvsec}.

\subsubsection{Zero-shot results on DENSE dataset}
To further assess robustness under domain shift, we evaluate the
EventScape-trained model on the unseen DENSE dataset~\cite{dense}.
As shown in Tab.~\ref{tab:dense}, SFE-VGGT achieves the second-best
performance across all three evaluation ranges, outperforming
EventDAM~\cite{eventdam} and the supervised RAMNet~\cite{ramnet} and
SRFNet~\cite{srfnet} baselines. Compared with EventDAM, the advantage
becomes more pronounced with depth, with \textbf{44.6\%} and
\textbf{54.6\%} lower errors at 20\,m and 30\,m, respectively. This wideningy
margin highlights the \textbf{robustness under a substantial domain shift} of the transferred geometry.


\begin{table}[t!]
\centering
\caption{
Zero-shot depth estimation on DENSE.\\
Mean absolute depth error is reported in meters.
}
 \vspace{-10pt}
\label{tab:dense}
\footnotesize
\setlength{\tabcolsep}{10pt}
\begin{tabular}{lcccc}
\toprule
Method & Input
& 10m $\downarrow$
& 20m $\downarrow$
& 30m $\downarrow$ \\
\midrule

RAMNet~\cite{ramnet}
& E+I
& 2.62 & 11.26 & 19.11 \\

SRFNet~\cite{srfnet}
& E+I
& 1.50 & 3.57 & 6.12 \\

\midrule

EventDAM~\cite{eventdam}
& E
& 1.20 & 2.60 & 5.18 \\

EventVGGT~\cite{eventvggt}
& E
& \best{0.54}
& \best{0.89}
& \best{1.33} \\

\textbf{SFE-VGGT}
& E
& \second{1.04}
& \second{1.44}
& \second{2.35} \\

\bottomrule
\end{tabular}
 \vspace{-10pt}
\end{table}

\subsubsection{Direct VGGT transfer on MVSEC Night1 scene}
This experiment examines whether pretrained VGGT~\cite{vggt} can be
directly applied to event-based depth estimation without event-domain
distillation. As shown in Tab.~\ref{tab:mvsec_zero}, SFE-VGGT
substantially improves over direct VGGT with event inputs, reducing the
error by \textbf{41.7\%}, \textbf{24.7\%}, and \textbf{15.1\%} at 10\,m, 20\,m, and
30\,m, respectively. It also achieves the best 10\,m result and remains
close to EventVGGT~\cite{eventvggt} at 20\,m. Overall, these results
\textbf{highlight the importance of event-domain distillation for
effectively exploiting VGGT's geometric priors}. Qualitative results
are shown in Fig.~\ref{fig:mvsec} (middle row).



\begin{table}[t!]
\centering
\caption{
Direct VGGT transfer analysis on MVSEC Night1. \\
Mean absolute depth error is reported in meters.
}
 \vspace{-10pt}
\label{tab:mvsec_zero}
\footnotesize
\setlength{\tabcolsep}{3.8pt}

\begin{tabular}{lc|ccc|ccc}
\toprule
Method & Input
& 10m $\downarrow$
& 20m $\downarrow$
& 30m $\downarrow$
& $\delta_1$ $\uparrow$
& $\delta_2$ $\uparrow$
& $\delta_3$ $\uparrow$ \\
\midrule

VGGT~\cite{vggt}
& E
& 2.42 & 2.71 & 3.44 & 0.42 & 0.69 & 0.84\\

VGGT~\cite{vggt}
& I
& 2.31 & 2.68 & 3.33 & 0.41 & \second{0.71} & \second{0.88} \\

\midrule

EventVGGT~\cite{eventvggt}
& E
& \second{1.67}
& \best{2.02}
& \best{2.61}
& \best{0.57} & \best{0.82} & \best{0.93} \\

\textbf{SFE-VGGT}
& E
& \best{1.41}
& \second{2.04}
& \second{2.92} 
& \second{0.48} & \second{0.71} & 0.83 \\

\bottomrule
\end{tabular}
 \vspace{-5pt}
\end{table}

\begin{table}[t!]
\centering
\caption{
Ablation of the proposed training objectives.
}
 \vspace{-10pt}
\label{tab:ablation_loss}
\scriptsize
\setlength{\tabcolsep}{3pt}

\begin{tabular*}{\columnwidth}{@{\extracolsep{\fill}}ccc|cccccc@{}}
\toprule
$\mathcal{L}_{DAFD}$
& $\mathcal{L}_{CWDD}$
& $\mathcal{L}_{CFRC}$
& 10m $\downarrow$
& 20m $\downarrow$
& 30m $\downarrow$
& $\delta_1$ $\uparrow$
& $\delta_2$ $\uparrow$
& $\delta_3$ $\uparrow$ \\

\midrule
             & $\checkmark$ &  & 0.620 & 0.951 & 1.337 & 0.740 & \second{0.880} & \best{0.948} \\
$\checkmark$ & $\checkmark$ &  & 0.622 & \second{0.947} & \second{1.327} & 0.739 & 0.875 & 0.942 \\
             & $\checkmark$ &  $\checkmark$ & \best{0.604} & 0.948 & 1.336 & \best{0.752} & \best{0.887} & \second{0.947} \\
$\checkmark$ & $\checkmark$ & $\checkmark$
             & \second{0.617} & \best{0.941} & \best{1.322} & \second{0.743} & \second{0.880} & \best{0.948} \\

\bottomrule
\end{tabular*}
\vspace{-10pt}
\end{table}

\subsection{Ablation Studies}
\label{sec:ablation}

We conduct all ablation experiments using the Town2 split of EventScape
for training and evaluate on the full EventScape test set. We examine
three aspects of the proposed framework: the contribution of each
training objective, the effectiveness of the key components introduced
within these objectives, and the influence of input sequence length.

\subsubsection{Effect of the Distillation Objectives}
Tab.~\ref{tab:ablation_loss} evaluates the contribution of the three
training objectives. CWDD is retained in all variants as the primary
depth-distillation objective, since removing it leads to severe
performance degradation. Adding DAFD mainly benefits the 20\,m and
30\,m ranges, suggesting that reliability-aware feature transfer is
more useful for \textbf{preserving geometry at larger depths}. In contrast,
CFRC gives the strongest 10\,m result and improves the stricter
$\delta$ metrics, indicating that \textbf{cross-frame relational supervision
is particularly effective for near-range depth consistency}. Combining
all three objectives yields the \textbf{most balanced performance
across both depth ranges and threshold-based metrics}, supporting the
complementary roles of DAFD and CFRC alongside CWDD. Since all variants
share the same VGGT initialization and retain CWDD under a reduced
training budget, the overall trends are more informative than small
differences in individual scores.

\subsubsection{Effect of Key Components in the Proposed Objectives}

Tab.~\ref{tab:ablation_design} evaluates the key mechanisms within
DAFD, CWDD, and CFRC. \textbf{Removing event-density weighting causes the most
consistent degradation}, particularly at the 20\,m and 30\,m ranges,
supporting its role in suppressing unreliable feature transfer from
sparsely observed regions. \textbf{Relative-confidence weighting also improves
performance across all ranges}, indicating the benefit of selectively
regulating depth supervision. \textbf{Removing RANSAC leads to a smaller but
consistent degradation}, showing that filtering unreliable
correspondences further improves the cross-frame constraint. Overall, the full configuration performs best across all depth ranges,
confirming that each reliability mechanism contributes to the proposed
distillation framework.


\begin{table}[t!]
\centering
\caption{
Ablation of key components within the proposed objectives.
}
 \vspace{-10pt}
\label{tab:ablation_design}
\footnotesize
\setlength{\tabcolsep}{4pt}

\begin{tabular}{lccc}
\toprule
Settings
& 10m $\downarrow$
& 20m $\downarrow$
& 30m $\downarrow$ \\

\midrule

w/o event-density weighting ($\mathcal{L}_{DAFD}$)
& 0.633 & 0.965 & 1.347 \\

w/o relative-confidence weighting ($\mathcal{L}_{CWDD}$)
& 0.631 & 0.954 & 1.334 \\

w/o RANSAC ($\mathcal{L}_{CFRC}$)
& 0.630 & 0.944 & 1.325 \\

Full model (all activated)
& \textbf{0.617} & \textbf{0.941} & \textbf{1.322} \\

\bottomrule
\end{tabular}
 \vspace{-10pt}
\end{table}









\subsubsection{Effect of Input Sequence Length}

Tab.~\ref{tab:ablation_seq} studies the influence of temporal context
on depth estimation. Increasing the sequence length consistently improves
the 20\,m and 30\,m errors as well as all $\delta$ metrics, while the
10\,m error shows only minor variation for shorter sequences. The
24-frame setting achieves the best performance across all reported
metrics, indicating that longer temporal windows provide richer
cross-frame geometric cues for the VGGT backbone. Overall,
\textbf{a longer input sequence leads to more reliable depth estimation},
with the clearest benefit appearing at larger depth ranges and in the
threshold-based metrics.


\begin{table}[t!]
\centering
\caption{
Effect of input sequence length.
}
 \vspace{-10pt}
\label{tab:ablation_seq}
\footnotesize
\setlength{\tabcolsep}{4pt}

\begin{tabular*}{\columnwidth}{@{\extracolsep{\fill}}c|cccccc@{}}
\toprule
Frames
& 10m $\downarrow$
& 20m $\downarrow$
& 30m $\downarrow$
& $\delta_1$ $\uparrow$
& $\delta_2$ $\uparrow$
& $\delta_3$ $\uparrow$ \\

\midrule

3  & 	0.666 &	1.034 &	1.510 &	0.693 &	0.834 &	0.911 \\
6  & 	0.681 &	1.025 &	1.471 &	0.706 &	0.847 &	0.921 \\
12 &    0.677 &	1.007 &	1.424 &	0.725 &	0.866 &	0.937 \\
24 &    \textbf{0.617} &	\textbf{0.941} &	\textbf{1.322} &	\textbf{0.743} &	\textbf{0.880} &	\textbf{0.943} \\

\bottomrule
\end{tabular*}
 \vspace{-6pt}
\end{table}






\section{Conclusion}
In this paper, we presented SFE-VGGT, a source-free distillation framework for transferring VGGT's geometric priors to event-based depth estimation without paired RGB observations or metric depth supervision. By using E2VID-reconstructed frames as teacher inputs, SFE-VGGT removes the need for paired RGB observations during distillation. To address the imperfect supervision introduced by event reconstruction, we propose Density-Aware Feature Distillation (DAFD), Confidence-Weighted Depth Distillation (CWDD), and Cross-Frame Relational Consistency (CFRC) for reliable feature, depth, and temporal geometric transfer. Extensive experiments on EventScape, MVSEC, and DENSE show that SFE-VGGT remains competitive with RGB-assisted distillation methods, performs particularly well under challenging lighting conditions, where conventional RGB observations are often degraded, and exhibits strong zero-shot generalization. These results demonstrate the feasibility of transferring geometric foundation-model priors to event-based modality without relying on synchronized RGB training data.

\smallskip
\noindent \textbf{Limitations and Future Work.}
The current framework relies on a frozen E2VID model to construct the teacher observations, and its performance can
therefore be influenced by the quality of the reconstructed frames.
While the proposed reliability-aware objectives reduce the impact of
uncertain reconstruction regions, the reconstruction module itself is
not adapted to the downstream geometric task. Future work will explore
joint optimization of the reconstruction and distillation framework,
allowing the reconstructed representation to better support the
extraction of geometric priors from VGGT while maintaining stable
training.


%
%
\bibliographystyle{ieeetr}
\bibliography{main}

\end{document}